\documentclass[runningheads]{llncs}
\usepackage{graphicx}
\usepackage{comment}
\usepackage{amsmath,amssymb} 
\usepackage{color}
\usepackage{soul}
\usepackage{xcolor}
\usepackage{algorithm}
\usepackage{algorithmic}
\usepackage{multirow}
\usepackage{colortbl}
\usepackage{cite}

\usepackage[width=122mm,left=12mm,paperwidth=146mm,height=193mm,top=12mm,paperheight=217mm]{geometry}

\definecolor{gd}{RGB}{206,206,206}
\definecolor{gl}{RGB}{240,240,240}

\begin{document}

\pagestyle{headings}
\mainmatter

\title{Self-supervised Feature Learning by Cross-modality and Cross-view Correspondences}

\titlerunning{}

\author{Longlong Jing \inst{1} \and
Yucheng Chen \inst{2} \and
Ling Zhang \inst{1} \and
Mingyi He \inst{2} \and
Yingli Tian \inst{1} \thanks{Corresponding author.}}

\authorrunning{L. Jing et al.}

\institute{City University of New York \and
Northwestern Polytechnical University}

\maketitle

\begin{abstract}
The success of supervised learning requires large-scale ground truth labels which are very expensive, time-consuming, or may need special skills to annotate. To address this issue, many self- or un-supervised methods are developed. Unlike most existing self-supervised methods to learn only 2D image features or only 3D point cloud features, this paper presents a novel and effective self-supervised learning approach to jointly learn both 2D image features and 3D point cloud features by exploiting cross-modality and cross-view correspondences without using any human annotated labels. Specifically, 2D image features of rendered images from different views are extracted by a 2D convolutional neural network, and 3D point cloud features are extracted by a graph convolution neural network. Two types of features are fed into a two-layer fully connected neural network to estimate the cross-modality correspondence. The three networks are jointly trained (i.e. cross-modality) by verifying whether two sampled data of different modalities belong to the same object, meanwhile, the 2D convolutional neural network is additionally optimized through minimizing intra-object distance while maximizing inter-object distance of rendered images in different views (i.e. cross-view). The effectiveness of the learned 2D and 3D features is evaluated by transferring them on five different tasks including multi-view 2D shape recognition, 3D shape recognition, multi-view 2D shape retrieval, 3D shape retrieval, and 3D part-segmentation. Extensive evaluations on all the five different tasks across different datasets demonstrate strong generalization and effectiveness of the learned 2D and 3D features by the proposed self-supervised method. 

\end{abstract}

\section{Introduction}

The deep convolutional neural networks for computer vision tasks (e.g. classification~\cite{qi2017pointnet,qi2017pointnet++}, detection~\cite{lang2019pointpillars}, segmentation~\cite{behley2019semantickitti}, etc.) are highly relied on large-scale labeled datasets \cite{kay2017kinetics,russakovsky2015imagenet}. Collecting and annotating the large-scale datasets are usually expensive and time-consuming. To facilitate 3D computer vision research, more and more 3D datasets such as mesh and point cloud data have been recently proposed. Compared to the annotation process of 2D image data, 3D point cloud data are especially harder to annotate and the cost is more expensive.

To learn features from unlabeled data, many self-/un-supervised learning methods are proposed for images, videos~\cite{goyal2019scaling,jing2019self,kolesnikov2019revisiting}, and 3D point cloud data~\cite{hassani2019unsupervised} by training deep neural networks to solve pretext tasks with automatically generated labels based on attributes of the data such as clustering images~\cite{caron2018deep,noroozi2018boosting}, playing image jigsaw~\cite{noroozi2016unsupervised}, predicting geometric transformation of images or videos~\cite{gidaris2018unsupervised,jing2018self}, image inpainting~\cite{pathak2016context}, reconstructing point cloud~\cite{yang2018foldingnet}, etc. The learned features through these processes are then used as pre-trained models for other tasks to overcome over-fitting and speed up convergence especially when training data is limited.

Recently self-supervised feature learning on 3D point cloud data attract more attention including auto-encoders-based methods~\cite{achlioptas2017learning,gadelha2018multiresolution,yang2018foldingnet,zhao20193d},   generative model-based methods~\cite{li2018point,sun2018pointgrow,wu2016learning}, and context-based pretext task method~\cite{hassani2019unsupervised,zhang2019unsupervised}. The auto-encoders-based and generative-based methods learn features by generating or reconstructing the point cloud data and have obtained very competitive performance on the 3D recognition benchmark \cite{yang2018foldingnet}. However, by optimizing the loss for generation or reconstruction tasks, these networks suffer from modeling low-level features and compromising their ability to capture high-level features from the point cloud data. 

\begin{figure}[tb]
\centering
\includegraphics[width = 0.9\textwidth]{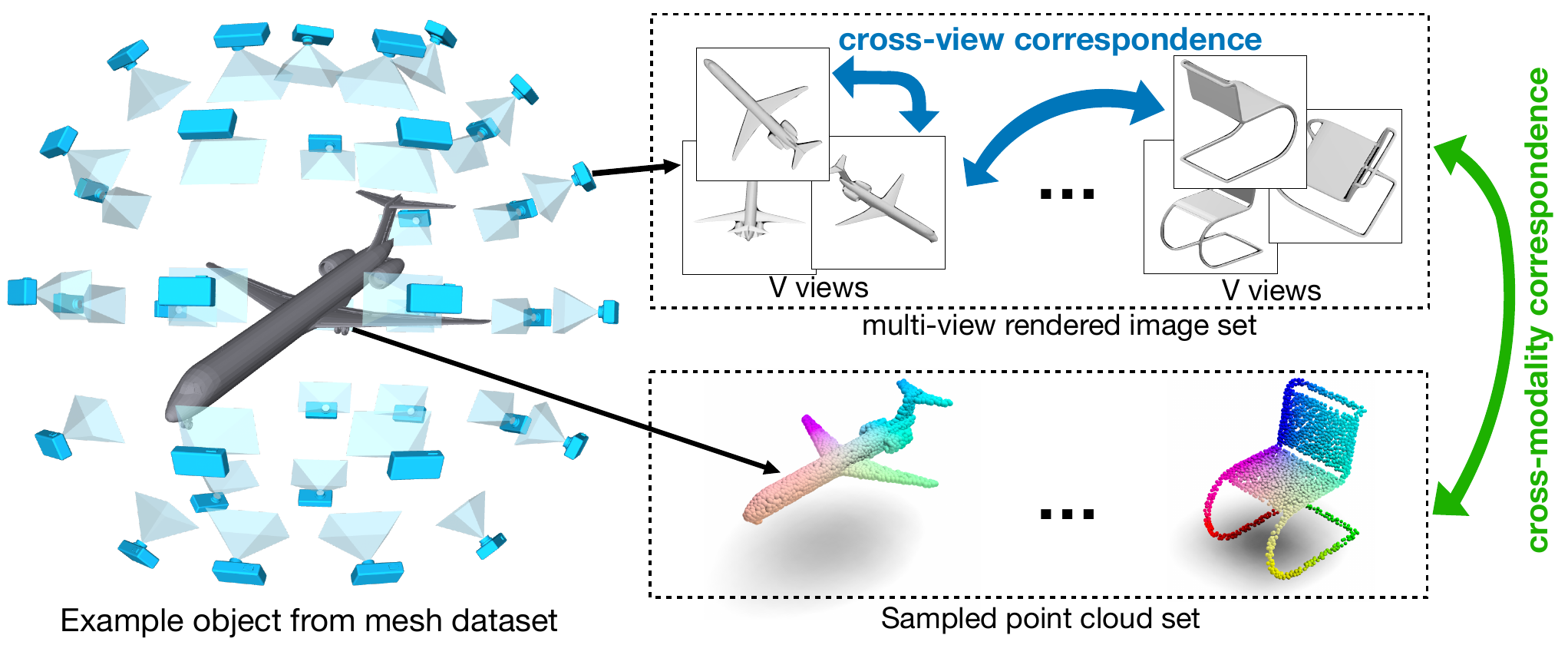}
\vspace{-10pt}
\caption{Training set generation. From 3D mesh datasets, multi-view rendered image set and sampled point cloud set are generated. The relations of the different data representations are employed as supervision signal (cross-view and cross-modality correspondences) to learn both 2D and 3D features without using any human annotated labels.}
\label{fig:motivation}
\vspace{-12pt}
\end{figure}

In this paper, as shown in Fig.~\ref{fig:motivation}, we propose a novel idea to explore how to use the abundant relations of different views and modalities of 3D data (e.g. mesh, point cloud, rendered shading images, rendered depth images, etc.) as supervision signal to learn both 2D and 3D features without using any human annotated labels. The main contributions of this paper are summarized as follows: 

\begin{itemize}
\item We design a new schema to jointly learn both 2D and 3D features through solving two parallel pre-defined pretext tasks: 1) Cross-modality task - to recognize whether two data in different modalities (3D point cloud and 2D image) belong to the same object; 2) Cross-view task - to minimize the distance of 2D image features in different views of the same object while maximizing the distance of 2D image features from different objects.

\item The discriminative 2D and 3D features learned by the self-supervised schema are used as pre-trained models for other down-stream tasks such as classification, retrieval, and 3D part segmentation, etc.

\item Extensive experiments on five different tasks (i.e. multi-view 2D shape recognition, 3D shape recognition, multi-view 2D shape retrieval, 3D shape retrieval, and 3D part-segmentation) demonstrate the effectiveness and generalization of the proposed framework. For the recognition tasks, our 2D and 3D models outperform the existing state-of-the-art unsupervised methods and achieve comparable performance as the supervised methods on the ModelNet40.

\end{itemize}

\section{Related Work}

\noindent\textbf{3D Point Cloud Understanding:} Various methods have been proposed for point cloud data understanding and they can be categorized into three types: hand-crafted methods \cite{chen2003visual,kazhdan2003rotation}  which use hand-designed feature extractors to model the geometric features; deep neural networks on regular 3D data  \cite{chang2015shapenet,feat1,feat2,volum4,volum1,volum5,su2015multi,su2018deeper,volum3} in which the network usually operates on multi-view rendered images \cite{su2015multi,su2018deeper} or volumetric voxelized data \cite{chang2015shapenet,volum4,volum5,volum1,volum3}; and deep neural networks on unordered 3D data in which the network operates directly on the unordered point cloud data \cite{lei2019spherical,li2019deepgcns,qi2017pointnet,qi2017pointnet++,thomas2019kpconv,wang2019dynamic,wu2019pointconv}. 3D point cloud data can be rendered into 2D images from different views to create multi-modality data. To utilize the multi-view images, Su \textit{et al.} proposed to tackle the 3D shape recognition by multi-view CNN operating on multiple 2D images that rendered from different views of the 3D data \cite{su2015multi}. To directly learn 3D features on unordered point cloud data, Qi \textit{et al.} proposed the milestone work PointNet by using a deep neural network to classify 3D shape data, and later this work was extended to many other networks \cite{li2019deepgcns,qi2017pointnet++,wang2019dynamic}. Wang \textit{et al.} proposed the EdgeConv with Multilayer Perceptron (MLP) to modal local features for each point from its k nearest neighbor (KNN) points. 

\noindent\textbf{2D Unsupervised Feature Learning:} Recently, many self-supervised learning methods (also known as unsupervised learning) have been proposed to learn features from unlabeled data \cite{caron2018deep,gidaris2018unsupervised,goyal2019scaling,jing2018self,jing2019self,kolesnikov2019revisiting,noroozi2018boosting,noroozi2016unsupervised,pathak2016context}. Usually, a pretext task is defined to train a network with automatically generated labels based on the attributes of the data. These methods fall into four groups: correspondence-based method (i.e. using the correspondence of two different modalities like visual and audio streams in videos as supervision signals) \cite{korbar2018cooperative}; context-based methods (i.e. using context structure or similarity of the data as supervision signals) \cite{caron2018deep,gidaris2018unsupervised,noroozi2016unsupervised,noroozi2018boosting}; generation-based methods (i.e. using the learned features in the process of generating images or videos such as Generative Adversarial Networks and Auto-encoder) \cite{pathak2016context}; and free semantic label-based methods (i.e. using the automatically generated labels by game engines or some traditional methods) \cite{pathak2017learning}. The 2D self-supervised learning has been well studied recently, and some methods have been successfully adapted to the 3D  self-supervised feature learning \cite{hassani2019unsupervised,sauder2019self,zhang2019unsupervised}.

\textbf{3D Self-supervised Feature Learning:} Several self-supervised learning methods have been proposed to model features from unlabeled 3D point cloud data  \cite{chen2003visual,gadelha2018multiresolution,kazhdan2003rotation,li2018point,sun2018pointgrow,wu2016learning,yang2018foldingnet,zhao20193d}. Most of these methods are auto-encoder based ~\cite{achlioptas2017learning,gadelha2018multiresolution,yang2018foldingnet,zhao20193d} to learn the features in the process of reconstructing the point cloud data or generative-based methods ~\cite{li2018point,sun2018pointgrow,thabet2019mortonnet,wu2016learning} to learn the features in the process of generating plausible point cloud data. Recently, a few work attempted to learn features by designing novel pretext tasks  \cite{hassani2019unsupervised,sauder2019self,zhang2019unsupervised}. Sauder \textit{et al.} proposed to learn features by recognizing the relative position of two segments from point cloud data \cite{sauder2019self}. Zhang \textit{et al.} proposed EdgeConv to learn features by verifying whether two segments are from the same object and then boosting the performance of a cluster task \cite{zhang2019unsupervised}. Hassani \textit{et al.} proposed a multi-task learning framework to learn features by optimizing three different tasks including clustering, prediction, and reconstruction \cite{hassani2019unsupervised}. However, all these methods only focus on learning one type of feature for 3D shape data while ignoring the inherent multi-modalities of different data representations. In this paper, we propose to learn two different types of features, 2D image features, and 3D point cloud features, by exploiting the correspondences of cross-modality and cross-view attributes of 3D data.

\section{Method}

Preparing 2D images in multiple views and 3D point cloud data from mesh objects is essential for our proposed self-supervised 2D and 3D feature learning.
The details of the data generation, the architecture of the framework, and model parameterization are introduced in the following sections.

\subsection{Data Generation}

As shown in Fig.~\ref{fig:motivation}, two types of training sets are generated from 3D object datasets, i.e., multi-view rendered image set and sampled point cloud set, for learning 2D and 3D features. 3D objects are typically represented in polygon meshes as collections of vertices, edges, and faces, etc. See Section~\ref{sec_param} for specific input samples for the framework.

\begin{figure}[tb]
\centering
\includegraphics[width=0.95\textwidth]{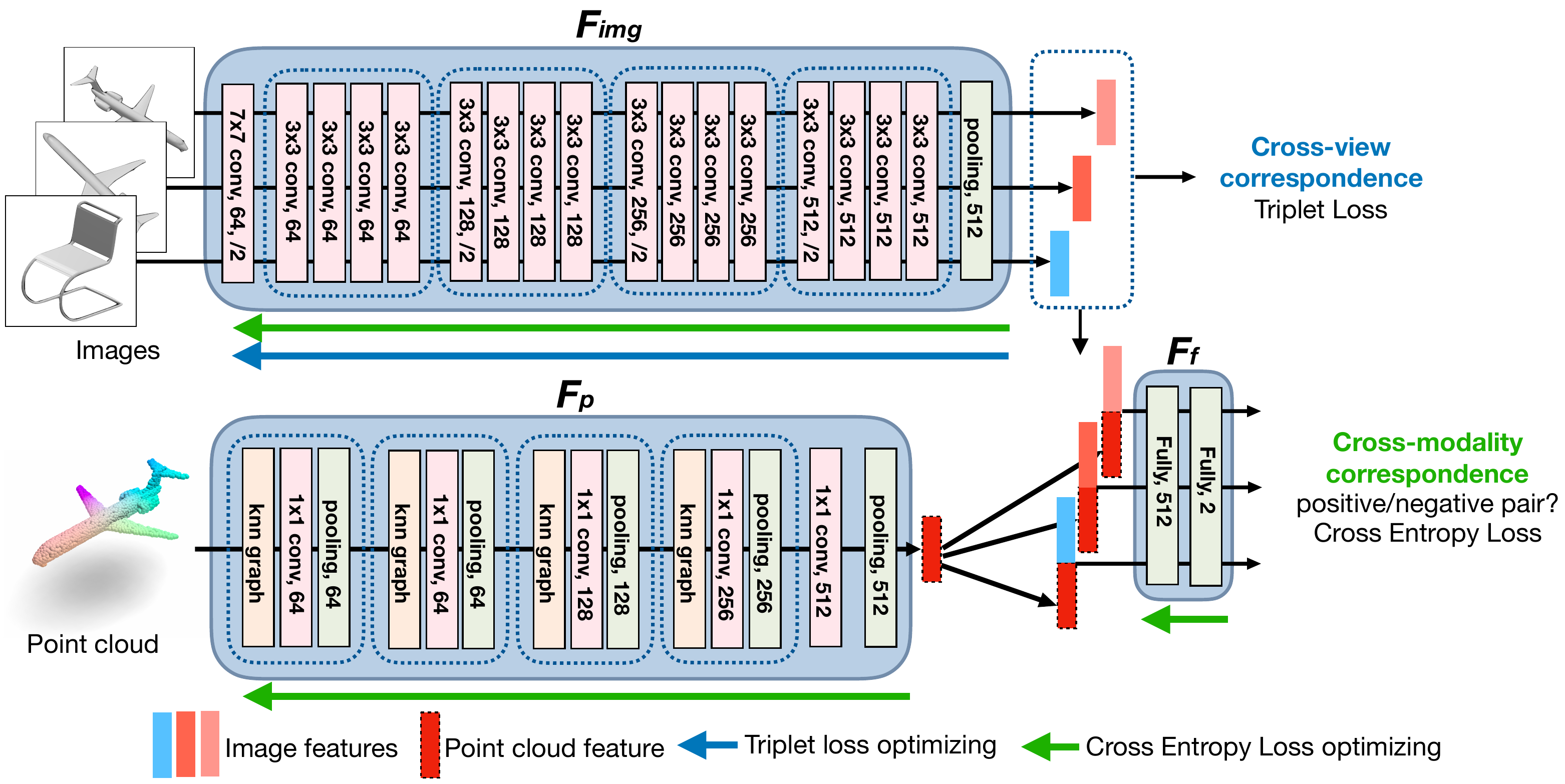}
\caption{The proposed framework for self-supervised 2D and 3D feature learning by cross-modality and cross-view correspondences. It consists of an image feature extracting 2DCNN ($F_{img}$) taking different rendered views, a graph neural network ($F_{p}$) taking unordered point cloud data, and a two-layer fully connected neural network ($F_{f}$) taking the concatenation of two types of features extracted by $F_{img}$ and $F_{p}$ to predict the cross-modality correspondence.  $F_{img}$, $F_{p}$, and $F_{f}$ are jointly trained (i.e. cross-modality, the blue solid arrow) by verifying whether two sampled data of different modalities belong to same object, meanwhile, $F_{img}$ is additionally optimized through minimizing intra-object distance while maximizing inter-object distance of rendered images in different views (i.e. cross-view, the green solid arrow).}
\label{fig:framework}
\vspace{-16pt}
\end{figure}

\textbf{Multi-view image generation:} Following~\cite{su2015multi}, the Phong reflection model \cite{phong1975illumination} is employed as the rendering engine to generate rendered images in different views from 3D polygon meshes. By given a 3D  polygon mesh $m$ from a 3D object set $M$, a spherical coordinate system is defined with the centroid of $m$ as the center for the system. The centroid for each $m$ is calculated as the average of all mesh face geometric centers of $m$, while the mesh face centers are weighted by the corresponding mesh face areas. To project $m$ to multi-view 2D planes, $V$ virtual cameras (viewpoints) around $m$ are randomly placed for each object along a sphere surface with radius $R$ (see Fig.~\ref{fig:motivation}). Each virtual camera is arranged by an azimuthal angle (randomly selected from 10 to 340 degrees) and a polar angle (randomly selected from 10 to 165 degrees) of the spherical coordinate system. All virtual cameras point toward the centroid of $m$, and one image is rendered form each camera. The intensities of pixels in the rendered images are determined by interpolating the reflected intensities of the polygon vertices. Due to the randomness of the sampled views, some parts of objects would be dark following the traditional settings if only one light source is placed during rendering. To avoid the problem, in our rendering process, two light sources are placed facing each other, while the mesh object is in between. The model shapes are uniformly scaled to fit into the perspective view. Note that $V$ images at different views are rendered for each 3D object, and up to two of the rendered images are used in each input training sample, and $v \leq V$ images are used in the testing phase.

\textbf{Point cloud sampling:}  Following ~\cite{qi2017pointnet}, we adopt the Farthest Point Sampling (FPS) algorithm to sample point clouds from each mesh object surface in the mesh datasets. Starting from a randomly chosen point, the next point is sampled in turn according to the average distance to all sampled points, that is, the farthest point. Each mesh object is uniformly sampled 2,048 points to keep the shape information of the object as much as possible. All sampled points are then normalized into a unit sphere.

\vspace{-5pt}
\subsection{Framework Architecture}\label{sec_Architecture}

As illustrated in Figure~\ref{fig:framework}, there are three networks in our framework: a 2DCNN ($F_{img}$) to extract 2D features from images cross different views, a graph neural network ($F_{p}$) to extract 3D features from unordered point cloud data, and a two-layer fully connected neural network $F_{f}$ to predict the cross-modality correspondence based on the two types of features extracted by $F_{img}$ and $F_{p}$. The three networks are jointly optimized by cross-modality correspondence, meanwhile, the network $F_{img}$ is optimized by cross-view correspondence (see details in Section~\ref{sec_param}).

The 2D image feature learning network ($F_{img}$) employs ResNet18~\cite{he2016deep} as the backbone network with four convolution blocks with a number of \{64, 128, 256, and 512\} $3 \times 3$ kernels. Each convolution block includes two convolution layers followed by a batch-normalization layer and a ReLU layer, except the first convolution block which consists of one convolution layer, one batch-normalization layer, and one max-pooling layer. A global average pooling layer, after the fourth convolution blocks, is used to obtain the global features for each image. Unless specifically pointed out, a 512-dimensional vector after the global average pooling layer is used for all our experiments.

The 3D point cloud feature learning network ($F_{p}$) employs dynamic graph convolutional neural network (DGCNN)~\cite{wang2019dynamic} as the backbone model due to its capability to model local structures of each point by dynamically constructed graphs and its good performance on classification and segmentation tasks. There are four EdgeConv layers and the number of kernels in each layer is $64$, $64$, $64$, and $128$, respectively. Each convolution graph consists of one KNN graph layer which builds the KNN graph for each point and two convolution layers. Each convolution layer is followed by a batch-normalization layer and a leaky ReLU layer. The EdgeConv layers aim to construct graphs over $k$ nearest neighbors calculated by KNN and the features for each point are calculated by an MLP over all the $k$ closest points. After the four EdgeConv blocks, a 512-dimension fully connected layer is used to extract per-point features for each point and then a max-pooling layer is employed to extract global features for each object.

The two-layer fully connected neural network $F_{f}$ is employed for cross-modality classification, which consists of a 256-dimensional fully connected layer and a 2-dimensional fully connected layer. Each feature vector feeding into $F_{f}$ is extracted by $F_{img}$ and $F_{p}$ and concatenated together as a 1024-dimension vector. The output of $F_{f}$ is a binary classification value. 

\subsection{Model Parameterization}\label{sec_param}

In our proposed self-supervised learning schema, two types of constraints are used as supervision signals to optimize the networks: cross-modality correspondence and cross-view correspondence. The cross-modality correspondence requires networks to learn modality-invariant features extracted from two different modalities $F_{img}$ and $F_{p}$, while the cross-view correspondence requires the subnetwork $F_{img}$ to capture semantic 2D image features to match objects from random views. We formulae the cross-modality task as a classification task and the cross-view task as a metric learning task. 

Let $\mathcal{D} = \{sample^{(1)}, ..., sample^{(N)}\}$ denotes training data of size $N$. The $i$-th input sample $sample^{(i)} = \{p^{(i)}, img_{1}^{(i)}, img_{2}^{(i)}, img_{3}^{(i)}, y_{1}^{(i)}, y_{2}^{(i)}, y_{3}^{(i)}\}$, where $p^{(i)}$ and $img_{1}^{(i)}$, $img_{2}^{(i)}$ represent the point cloud and two different rendered views generated from the same 3D mesh object respectively, and $img_{3}^{(i)}$ is an image rendered from a different object. The labels $y_{j}^{(i)} \in \{0, 1\}$ indicates whether the point cloud $p^{(i)}$ and the rendered image $img_{j}^{(i)}$ are from same object where $1$ for same object and $0$ for different objects. Note that $img_{1}^{(i)}$ and $img_{2}^{(i)}$ are randomly selected in $V$ rendered views from a 3D mesh object same as the sampled point cloud $p^{(i)}$, while $img_{3}^{(i)}$ is from a different one.

\textbf{Cross-view correspondence:} The objective of the cross-view task is to train the network $F_{img}$ to learn view invariant features from rendered images. When an object observed from different views, the visible parts may look differently, however, the semantic features for images in different views should be similar. Therefore, triplet loss \cite{schroff2015facenet} is employed here to train the network to minimize distance of features of positive pairs (i.e. from same object) and maximize distance of features of negative pairs (i.e. from different objects):
\begin{align}
\begin{split}
L_{triplet} = & max(\Vert F_{img}(img_{1}^{(i)}) - F_{img}(img_{2}^{(i)}) \Vert^{2} \\
&- \Vert F_{img}(img_{1}^{(i)}) - F_{img}(img_{3}^{(i)}) \Vert^2 +\alpha, 0),
\end{split}
\end{align}
\noindent
where the triple samples $img_{1}^{(i)}$, $img_{2}^{(i)}$ and $img_{3}^{(i)}$ correspond to anchor, positive and negative rendered images, $\alpha$ is the margin hyper-parameter to control the differences of intra- and inter- objects.

\textbf{Cross-modality correspondence:} The cross-modality learning is modeled as a binary classification task by employing the cross-entropy loss to optimize all the three networks. After obtaining image features by $F_{img}$ from rendered images and point cloud features by $F_{p}$ from point clouds, the network $F_{f}$ predicts whether the two input data of different modalities are from same object by discovering the high-level modality invariant features. The positive samples are the point cloud and image pairs from same 3D mesh object, while the negative samples are from different objects. The loss function for jointly optimizing networks $F_{img}$, $F_{p}$, and $F_{f}$ is:
\begin{align}
\begin{split}
L_{cross} = & - \sum_{j=1}^{3}(y_{j}^{(i)}\log(F_{f}(F_{img}(img_{j}^{(i)}), F_{p}(p^{(i)}))) \\
&+ (1-y_{j}^{(i)})\log(1-F_{f}(F_{img}(img_{j}^{(i)}),  F_{p}(p^{(i)})))),
\end{split}
\end{align}

\noindent
The input features of $F_{f}$ are extracted by $F_{img}$ and $F_{p}$, and $F_{f}$ learns the correlation of the features extracted from two different data modalities.

When jointly train the three networks, a linear weighted combination of the loss functions $L_{triplet}$ and $L_{cross}$ are employed to optimize the whole framework. The final self-learning loss is combined as:
\begin{align}\label{eq:loss_combined}
L_{self} = L_{triplet} + \beta L_{cross},
\end{align}

\noindent
where $\beta$ is the weight for the cross-modality loss.

\begin{algorithm}[!t]
    \scriptsize
    \caption{The proposed self-supervised feature learning algorithm.}
    \label{alg:training}
    \begin{algorithmic}
        \STATE mini-batch size: $B$; 2D image features:$fi$; 3D point cloud features: $fp$; binary prediction: $\hat{y}$;
        
        \FOR{\textbf{all} sampled mini-batch $\{sample^{(b)}\}_{b=1}^{B}$}
            \STATE \textbf{for all} $b\in \{1, \ldots, B\}$ \textbf{do}
                \STATE $~~~~$ \textcolor{gray}{\# feature extraction}
                \STATE $~~~~$ $fi_{1}^{(b)} = F_{img}(img_{1}^{(b)})$; $fi_{2}^{(b)} = F_{img}(img_{2}^{(b)})$; $fi_{3}^{(b)} = F_{img}(img_{3}^{(b)})$;
                \STATE $~~~~$ $fp^{(b)} = F_{p}(p^{(b)})$;
                \STATE $~~~~$ \textcolor{gray}{\# classification prediction}
                \STATE $~~~~$ $\hat{y}_{1}^{(b)} = F_{f}(fi_{1}^{(b)}, fp^{(b)})$; $\hat{y}_{2}^{(b)} = F_{f}(fi_{2}^{(b)}, fp^{(b)})$; $\hat{y}_{3}^{(b)} = F_{f}(fi_{3}^{(b)}, fp^{(b)})$;
                \STATE $~~~~$ \textcolor{gray}{\# loss calculation}
                \STATE $~~~~$ $\mathcal{L}_{triplet}^{(b)} = max(\Vert fi_{1}^{(b)} - fi_{2}^{(b)}) \Vert^{2} - \Vert fi_{1}^{(b)} - fi_{3}^{(b)} \Vert^2 +\alpha, 0)$
                \STATE $~~~~$ $L_{cross}^{(b)} = - \sum_{j=1}^{3}(y_{j}^{(b)}\log(\hat{y}_{j}^{(b)}) + (1-y_{j}^{(b)})\log(1-\hat{y}_{j}^{(b)}))$ 
                \STATE $~~~~$ $L_{self}^{(b)} = L_{triplet}^{(b)} + \beta L_{cross}^{(b)}$
            \STATE \textbf{end for}
            \STATE $\mathcal{L} = \frac{1}{B} \sum_{b=1}^B L_{self}^{(b)}$
            \STATE update networks $F_{img}$, $F_{p}$ and $F_{f}$ to minimize $\mathcal{L}$
        \ENDFOR
        \STATE \textbf{return} pre-trained networks $F_{img}$ and $F_{p}$
    \end{algorithmic}
\end{algorithm}

The details of the joint training process are illustrated in \textit{Algorithm~\ref{alg:training}}. After the jointly training finished, two networks $F_{img}$ and $F_{p}$ are obtained as pre-trained models for two different modalities. The joint training enables the two feature extractors to learn more discriminative and robust features cross different data domains.

\section{Experimental Results}

\subsection{Experimental Setup}

\textbf{Self-supervised learning:} The proposed framework is optimized end-to-end using the SGD optimizer with an initial learning rate of $0.001$, the moment of $0.9$, and weight decay of $0.0005$. The learning rate decreases by $90\%$ every $40,000$ iteration. The networks for self-supervised learning are trained on the ModelNet40 dataset for $120,000$ iterations using a mini-batch size of $32$. To learn more robust features, data augmentation is applied to both images and point clouds. The images are randomly cropped and randomly flipped with $50$\% probability in the horizontal direction, while the point clouds are randomly rotated between [$0$, $2\pi$] degrees along the up-axis, randomly jittered the position of each point by Gaussian noise with zero mean and $0.02$ standard deviation. The rendering views $V$ is 180 for each 3D mesh object in the dataset. During the testing, we randomly select 2D-2D and 2D-3D testing pairs from the test split of ModelNet40 and ModelNet10. The amount of two types of pairs is ten times the test split including half positive pairs and half negative pairs.

\noindent \textbf{Evaluation of learned 2D and 3D features:} To evaluate the effectiveness and generalization of the learned 2D and 3D features by the proposed self-supervised learning schema, five different tasks are designed as follows. For the multi-view 2D shape recognition and 3D shape recognition tasks, the image and point cloud features are extracted by two pre-trained networks $F_{img}$ and $F_{p}$, then trained on corresponding SVMs with one class linear kernel, respectively. For the 3D part segmentation task, additional fully connected layers are added on top of the pre-trained $F_{p}$ and then fine-tuned on the ShapeNet~\cite{chang2015shapenet} dataset. The network is optimized with Adam optimizer~\cite{kingma2014adam} using an initial learning rate of $0.003$ and decreased by $90$\% every 20 epochs. For the 2D and 3D shape retrieval tasks, Euclidean distance over the global features of two objects is used as a metric to measure the similarity of two objects.

\noindent \textbf{Datasets:} All the experiments are conducted on two 3D object benchmarks: ModelNet40~\cite{wu20153d} and ShapeNet~\cite{chang2015shapenet}. The ModelNet40 dataset contains $12,311$ meshed models covering $40$ classes, of which $9,843$ are used for training and $2,468$ for testing. The ModelNet40 is used to train our proposed self-supervised learning framework as well as for the evaluation tasks of multi-view 2D shape recognition and 3D shape recognition. The ModelNet10, a subset of ModelNet40, is also used as a testing set, which contains $10$ classes. The ShapeNet contains 16 object categories including $12,137$ models for training and $2,874$ for testing and it is employed to evaluate the task of 3D part segmentation. In all experiments, $2,048$ points are sampled for each 3D mesh object as the input point cloud data.

\subsection{Cross-modality and Cross-view Correspondence Evaluation}\label{sec_pretext}

A straightforward evaluation of the effectiveness of our proposed self-supervised learning framework is to recognize the cross-modality and cross-view correspondence with ModelNet40 and ModelNet10 datasets. Table~\ref{tab:SelfResult} reports the cross-modality recognition accuracy and cross-view feature Euclidean distance of testing image pairs.

\begin{table}[ht]
\caption{Performance on pretext task: cross-modality recognition accuracy and cross-view feature distance analysis. CM indicates network training with cross-modality correspondence. CV indicates network training with cross-view correspondence. mPD indicates mean Pair Distance with corresponding standard deviation in brackets. }
\begin{center}
\scalebox{0.96}{
\begin{tabular}{l|l|c|l|c|c}
\hline
\multirow{2}*{Testing set} & \multirow{2}*{Network}   & Cross-modality & \multirow{2}*{Network} &  \multirow{2}*{Positive mPD} & \multirow{2}*{Negative mPD}\\
 &    & Accuracy (\%)  &  & & \\
\hline
\multirow{2}*{ModelNet40}  & $F_{p}$-CM          & $93.5$  & $F_{img}$-CM   & $6.43$ (2.38)   & $12.07$ (3.46) \\
  & $F_{p}$-CM-CV    & $91.8$  & $F_{img}$-CM-CV   & $2.56$ (0.56)   & $4.33$ (1.37)  \\
\hline
\multirow{2}*{ModelNet10}  & $F_{p}$-CM          & $92.0$  & $F_{img}$-CM  & $6.83$ (2.36)   &$11.29$ (3.15)  \\
  & $F_{p}$-CM-CV    & $91.5$  & $F_{img}$-CM-CV  & $2.571$ (0.52)  &$4.304$ (1.06)  \\
\hline
\end{tabular}
}
\end{center}
\label{tab:SelfResult}
\vspace{-18pt}
\end{table}

For the cross-modality recognition task, our networks accomplish over $90$\% accuracy (see the third column) which shows that the self-supervised learning successfully learns modality invariant features. For the cross-view correspondence recognition, the margins between the mean distance of positive pairs and that of negative pairs are very large which demonstrates that the networks indeed learn the view-invariant features. When the networks trained jointly with cross-modality and cross-view correspondence, although the performance of cross-modality recognition decreases a little bit, the standard deviations for the distances of both positives and negatives are significantly improved (see rows 2 and 4) which validate that the cross-view correspondence enforces the learning of view-invariant features. 

\begin{figure}
\vspace{-10pt}
\centering
\includegraphics[width=0.95\textwidth]{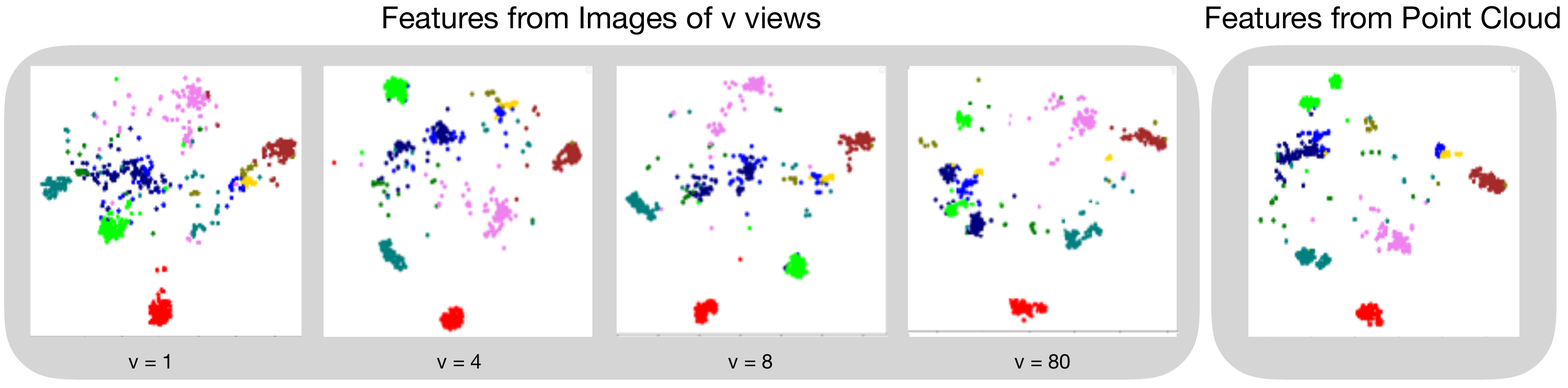}
\caption{Visualization of 2D and 3D features of the top 10 object categories (ten different colors) on the ModelNet40 test set. When more views in the testing phase are used to represent 3D objects, the distribution of 2D image features for different category objects is more discriminative, and is more similar to 3D point cloud feature distribution.}
\label{fig:tsne}
\end{figure}

One common problem of self-supervised learning is that the network can easily learn trivial features (e.g. corners, edges, or other low-level features) instead of high-level semantic features. To further analyze the features extracted by $F_{img}$ and $F_{p}$, we use T-distributed Stochastic Neighbor Embedding (TSNE)~\cite{maaten2008visualizing} to visualize the learned 2D and 3D features of the top 10 object categories in ten different colors on ModelNet40 as shown in Fig.~\ref{fig:tsne}. Each point indicates one feature that is max-pooled from $v$ extracted features of $v$ views. In the feature space, the features belong to the same class are closer than the features from different object classes, which show that the network indeed can learn high-level semantic features.

\subsection{Transfer to 2D and 3D shape recognition tasks}\label{sec_recognition}

Our proposed framework effectively learns both 2D and 3D features and achieves high performance on the pretext task. Here, we further evaluate the learned 2D and 3D features (i.e. $F_{img}$ and $F_{p}$) as pre-trained models on other down-stream supervised tasks: 2D and 3D shape recognition on ModelNet40 dataset. Two linear SVM classifiers are trained based on the extracted 2D and 3D features by $F_{img}$ and $F_{p}$, respectively. Same as in subsection~\ref{sec_pretext}, each extracted feature for 2D recognition task is max pooled from $v$ extracted features of $v$ random views, except when $v = 1$.

\begin{table}[ht]
    \caption{The performance of using the self-supervised learned models as feature extractors on the 2D and 3D shape recognition tasks on the ModelNet40 dataset. Both 2D and 3D shape recognition tasks are benefited from jointly training with cross-view and cross-modality correspondences. When multiple views (\#Views = 12, 36, or 80) are available for testing, the performance of 2D shape recognition is significantly improved.}
    \begin{center}
     \scalebox{1.0}{
        \begin{tabular}{l|c|c|l|c}
            \hline
            \multirow{2}*{Network}   & testing & 2D recognition  & \multirow{2}*{Network} & 3D recognition \\
            &  \#views & Accuracy (\%)  &  & Accuracy (\%) \\
            \hline
            $F_{img}$-CM               & 1 & $66.1$    &   $F_{p}$-CM     & $87.5$  \\
            $F_{img}$-CM-CV     & 1 & $72.5$  (+6.4)     &   $F_{p}$-CM-CV     & $\textbf{89.8}$ (+2.3)   \\
            $F_{img}$-CM-CV    & 12 & $87.3$ (+21.1)            & &  \\
            $F_{img}$-CM-CV    & 36 & $88.7$ (+22.6)          & &   \\
            $F_{img}$-CM-CV   & 80 & $\textbf{89.3}$ (+23.2)  & &  \\
            \hline
        \end{tabular}
        }
    \end{center}
    \label{tab:Classification}
    \vspace{-15pt}
\end{table}

As shown in Table~\ref{tab:Classification}, both the pre-trained $F_{img}$ and $F_{p}$ can achieve high accuracy on the 2D and 3D shape recognition tasks ($89.3$\% and $89.8$\%) to recognize $40$ object categories on ModelNet40 dataset which show that the two networks learn discriminative semantic features through the self-supervised learning process. When only trained with cross-modality correspondence, the 3D features learned by $F_{p}$-CM achieve $87.5$\% accuracy while the performance of 2D features by $F_{img}$-CM is only $66.1$\%. The joint training of cross-view and cross-modality correspondence significantly improves the performance of $F_{img}$ on 2D recognition ($+6.4\%$) and $F_{p}$ on 3D recognition ($+2.3\%$). The accuracy of 2D recognition is further boosted by more discriminative image features max pooled from multi-testing-view features, achieving 89.3\% with 80 views from each data.

\subsection{Transfer to 2D and 3D shape retrieval tasks}

{
To evaluate the generalization ability of the learned features, we further evaluate both 2D and 3D features extracted by $F_{img}$ and $F_{p}$ on shape retrieval tasks on the ModelNet40 dataset and Top-K accuracy are reported in Table~\ref{tab:retrieval}.

\begin{table}[ht]
\caption{Performance of the learned 2D and 3D features on the 2D and 3D shape retrieval tasks on ModelNet40 dataset. When only using $1$ view for each image, our self-supervised model $F_{img}$-CM-CV outperforms the ImageNet pre-trained model.}
\begin{center}
\scalebox{0.98}{
\begin{tabular}{l|c|c|c|c|c|c}
\hline
Network &Views &  Top1 (\%)  & Top5 (\%)  & Top10 (\%) & Top20 (\%) & Top50 (\%) \\
\hline
$F_{p}$-CM  &--- &$82.9$  &$94.2$  &$96.0$  &$97.8$  &$98.7$  \\
$F_{p}$-CM-CV &--- &$84.0$  &$94.3$  &$96.5$  &$97.9$  &$98.8$\\
\hline
ImageNet \cite{he2016deep}  &$1$  &$61.4$  & $82.1$ &$88.5$  &$93.5$ &$97.3$ \\
$F_{img}$-CM  &$1$  &$54.6$  &$79.2$  &$87.1$  &$93.0$  &$97.1$\\
$F_{img}$-CM-CV  &$1$   &$\textbf{66.9}$  &$\textbf{85.8}$  &$\textbf{91.1}$  &$\textbf{94.5}$  &$\textbf{97.9}$\\
\hline
ImageNet \cite{he2016deep}  &$12$  &$83.6$  & $94.7$ &$96.7$  &$98.0$ &$99.3$ \\
$F_{img}$-CM  &$12$  &$75.5$  &$91.2$  &$95.0$  &$97.2$  &$98.4$\\
$F_{img}$-CM-CV  &$12$  &$83.5$  &$94.2$  &$96.2$  &$97.5$  &$99.0$\\
\hline
ImageNet \cite{he2016deep} &$80$ &$87.6$  & $95.7$ &$97.4$  &$98.7$ &$99.4$  \\
$F_{img}$-CM  &$80$  &$82.7$  & $93.9$ &$96.4$  &$97.6$ &$98.9$  \\
$F_{img}$-CM-CV  &$80$  &$84.7$  &$94.7$  &$96.6$  &$97.9$  &$99.2$\\
\hline
\end{tabular}
}
\end{center}
\label{tab:retrieval}
\vspace{-10pt}
\end{table}

Since no other self-supervised learning methods for point cloud or multi-view images have reported performance on this task, we directly compare with ImageNet pre-trained models on the retrieval task. The 3D network $F_{p}$-CM and $F_{p}$-CM-CV can accomplish the retrieval task with high accuracy. As for the 2D network $F_{img}$, the performance is significantly improved when more views are used to represent each object. When only using $1$ view for each image, our self-supervised model $F_{img}$-CM-CV outperforms the ImageNet pre-trained model. When more views (12 or 80) are available, our model achieves comparable performance with the supervised model which is pre-trained on the ImageNet dataset.}

\subsection{Transfer to 3D part segmentation task}

{
To further verify the quality of 3D features learning by the pre-trained $F_{p}$ for point cloud data,  we conduct the transfer learning on the 3D part segmentation task with the ShapeNet dataset. To adapt $F_{p}$ on the 3D part segmentation task, four fully connected layers are added on the top of $F_{p}$, and the output from all the four blocks and the global features are used to predict the pixel-wise labels. Three sets of experiments are studied: (1) Only update the four newly added layers with frozen $F_p$, (2) $F_p$ and newly added layers are randomly initialized and supervised trained from scratch~\cite{qi2017pointnet}, (3) The learned features by $F_p$ are used as pre-trained models and all the layers are fine-tuned (unfrozen). The extensive studies of train/fine-tune strategies with different amounts of training data on the ShapeNet dataset for the 3D part segmentation are shown in Table~\ref{tab:partseg}.

\begin{table}[ht]
\vspace{-8pt}
\caption{The performance of the three types of settings on different amount of data from the ShapeNet dataset. $F_{p}$ with parameter-unfrozen setup outperforms the supervised method. When only a very small amount of data ($2\%$) is available for training, all our models outperform the supervised model.}
\begin{center}
\scalebox{0.99}{
\begin{tabular}{l|c|c|c|c}
\hline
\multirow{2}*{Network}   & Training  & Overall  & Class  & Instance  \\
   &  data &  Accuracy (\%) &  mIOU (\%) &  mIOU (\%) \\

\hline
$F_{p}$-CM-Frozen     &100\%  & $90.8$     & $71.2$    & $78.6$ \\
$F_{p}$-CM-CV-Frozen     &100\%  & $92.1$     & $74.7$    & $80.8$ \\
\cellcolor{gl}$F_{p}$-Supervised \cite{qi2017pointnet} &\cellcolor{gl}100\%  &\cellcolor{gl}$92.9$      &\cellcolor{gl}$77.6$   &\cellcolor{gl}$83.0$ \\  
$F_{p}$-CM-Unfrozen   &100\%  & $93.2$ ({+0.3})    & $78.1$ (+0.5) & $83.4$ (+0.4) \\
$F_{p}$-CM-CV-Unfrozen   &100\%  & $\textbf{93.4}$ (+0.5)    & $\textbf{79.1}$ (+1.5) & $\textbf{83.7}$ (+0.7) \\
\hline
$F_{p}$-CM-Frozen     &20\%  & $89.2$    & $65.6$    & $75.4$ \\
$F_{p}$-CM-CV-Frozen    &20\%  & $90.7$    & $68.5$    & $77.8$ \\
\cellcolor{gl}$F_{p}$-Supervised \cite{qi2017pointnet} &\cellcolor{gl}20\%  &\cellcolor{gl}$90.9$  &\cellcolor{gl}$69.9$   &\cellcolor{gl}$79.1$ \\ 
$F_{p}$-CM-Unfrozen   &20\%  & $91.3$ (+0.4)  & $70.9$ (+1.0)    & $80.0$ (+0.9) \\
$F_{p}$-CM-CV-Unfrozen   &20\%  & $\textbf{91.8}$ (+0.9)  & $\textbf{72.2}$ (+2.3)    & $\textbf{80.3}$ (+1.2) \\
\hline
$F_{p}$-CM-Frozen    &2\%  & $85.1$  (+0.1)   & $57.1$ (+0.9)    & $69.2$ (+0.2)\\
$F_{p}$-CM-CV-Frozen     &2\%  & $87.3$ (+2.3)   & $58.4$ (+2.2)   & $72.1$ (+3.1)\\
\cellcolor{gl}$F_{p}$-Supervised \cite{qi2017pointnet} &\cellcolor{gl}2\%  &\cellcolor{gl}${85.0}$ &\cellcolor{gl}${56.2}$     &\cellcolor{gl}${69.0}$ \\
$F_{p}$-CM-Unfrozen   &2\% & $87.2$ (+2.2) & $60.6$ (+4.4) & $72.6$ (+3.6)\\
$F_{p}$-CM-CV-Unfrozen &2\% & $\textbf{88.3 (+3.3)}$ & $\textbf{60.7 (+4.5)}$ & $\textbf{74.0 (+5.0)}$\\
\hline
\end{tabular}
}
\end{center}
\label{tab:partseg}
\vspace{-15pt}
\end{table}

As shown in Table~\ref{tab:partseg}, training with cross-view correspondence can improve the ability of $F_{p}$ to recognize object parts. When $100\%$ of the training data are available, even without updating the parameters of $F_p$ on the new task, it still achieves $80.8\%$ instance mIOU which is only $2.2\%$ lower than the supervised model. It validates that $F_p$ can learn semantic features from the proposed pretext task and transfer them across datasets and tasks. When the full network is initialized with the pre-training weights and further fine-tuned, the instance mIOU improves by $0.4\%$ and the class mIOU improves by $0.5\%$ showing that the learned weights for $F_p$ from self-supervised pretext task can be served as a good starting point for the optimization. When using only $20\%$ data, the parameter-unfrozen setup can significantly ($+2.3\%$ on class mIOU, and $+1.2\%$ on instance mIOU) boost up the performance than the supervised setup. When using only $2\%$ of the data, the performance of both parameter-frozen setup ($+2.2\%$ on class mIOU, and $+3.1\%$ on instance mIOU) and parameter-unfrozen setup ($+4.5\%$ on class mIOU, and $+5.0\%$ on instance mIOU) are better than the supervised setup. Our pre-trained $F_{p}$ performs well when fine-tuned on small-scale 3D shape datasets.}

\subsection{Comparison with the State-of-the-Art methods}\label{Comparison}

In this section, we further compare our pre-trained $F_{img}$ and $F_{p}$ with the state-of-the-art methods for 2D and 3D shape recognition on ModelNet40 dataset including 2D multi-view methods~\cite{donahue2014decaf,sanchez2013image,su2015multi} and 3D methods of both unsupervised learning models~\cite{achlioptas2017learning,chen2003visual,gadelha2018multiresolution,girdhar2016learning,hassani2019unsupervised,kazhdan2003rotation,sharma2016vconv,wu2016learning,yang2018foldingnet,zhao20193d} and supervised learning models~\cite{gadelha2018multiresolution,hua2018pointwise,klokov2017escape,li2018point,qi2017pointnet++,shen2018mining,wang2018local,wang2019dynamic}. The setups of our models are same as in subsection~\ref{sec_recognition}. The two types of comparisons are shown in Table~\ref{tab:2D} and Table~\ref{tab:3D}.

\begin{table}[ht]
    \caption{The comparison with 2D state-of-the-art methods for multi-view shape recognition on ModelNet40 dataset.}
    \begin{center}
    \scalebox{1.0}{
        \begin{tabular}{l|c|c|c|c}
            \hline
            \multirow{2}*{Network}   &  \multicolumn{2}{c|}{Training} &  Testing & Classification \\
            \cline{2-4}
            & Pre-train & Fine-tune  &  \#views & Accuracy (\%) \\
            \hline
            Fisher Vector~\cite{sanchez2013image}     & -  & ModelNet40     & 1 & $78.8$ \\
            DeCAF~\cite{donahue2014decaf}     & ImageNet1K  & -        & 1 & $83.0$ \\
            DeCAF, 12$\times$~\cite{donahue2014decaf}    & ImageNet1K  & ModelNet40     & 12 & $88.6$ \\
            MVCNN, 12$\times$~\cite{su2015multi}     & ImageNet1K  & -        & 12 & $88.6$ \\
            MVCNN, 12$\times$~\cite{su2015multi}     & ImageNet1K  & ModelNet40       & 12 & $89.9$ \\
            MVCNN, 80$\times$~\cite{su2015multi}     & ImageNet1K  & -        & 80 & $84.3$ \\
            MVCNN, 80$\times$~\cite{su2015multi}     & ImageNet1K  & ModelNet40  & 80 & $\textbf{90.1}$ \\
            \hline
            $F_{img}$-CM     & ModelNet40  & -       & 1 & $66.1$ \\
            $F_{img}$-CM-CV  & ModelNet40  & -         & 1 & $72.5$  \\
            $F_{img}$-CM-CV  & ModelNet40  & -         & 12 & $87.3$  \\
            $F_{img}$-CM-CV  & ModelNet40  & -         & 80 & $\textbf{89.3}$  \\
            \hline
        \end{tabular}
        }
    \end{center}
    \label{tab:2D}
\end{table}

The models for comparison in Table~\ref{tab:2D} are hand-crafted model~\cite{sanchez2013image} and supervised feature learning models~\cite{donahue2014decaf}, \cite{su2015multi}. Note that DeCAF and MVCNN are pre-trained on ImageNet1K~\cite{deng2009imagenet} and then fine-tuned on ModelNet40 datasets which demand large-scale labeled data for pre-training. When using the same number of views (\#views = $80$), the performance of our model is comparable to the state-of-the-art supervised method.

\begin{table}[ht]
    \caption{The comparison with the state-of-the-art methods for 3D point cloud shape recognition on ModelNet40 dataset.}
    \begin{center}
    \scalebox{1.0}{
        \begin{tabular}{l|c|l|c}
            \hline
            \multicolumn{2}{c|}{Unsupervised feature learning} & \multicolumn{2}{c}{Supervised feature learning} \\
            \hline
            \multirow{2}*{Network}   & Classification & \multirow{2}*{Network}   & Classification \\
              & Accuracy (\%) &   & Accuracy (\%)\\
            
            \hline
            SPH~\cite{kazhdan2003rotation}  & $68.2$  & PointNet~\cite{li2018point}  & $89.2$ \\
            LFD~\cite{chen2003visual}  & $75.5$  & PointNet++~\cite{qi2017pointnet++}  & $90.7$ \\
            T-L Network~\cite{girdhar2016learning}  & $74.4$  & PointCNN~\cite{hua2018pointwise}  & $86.1$ \\
            VConv-DAE~\cite{sharma2016vconv}  & $75.5$  & DGCNN~\cite{wang2019dynamic}  & $\textbf{92.2}$ \\
            3D-GAN~\cite{wu2016learning}  & $83.3$  & KCNet~\cite{shen2018mining}  & $91.0$ \\
            Latent-GAN~\cite{achlioptas2017learning}  & $85.7$  & KDNet~\cite{klokov2017escape}  & $91.8$ \\
            MRTNet-VAE~\cite{gadelha2018multiresolution}  & $86.4$  & MRTNet~\cite{gadelha2018multiresolution}  & $91.7$ \\
            Contrast-Cluster~\cite{zhang2019unsupervised} &$86.8$ &SpecGCN~\cite{wang2018local} &$91.5$ \\
            FoldingNet~\cite{yang2018foldingnet}  & $88.4$  &   &  \\
            PointCapsNet~\cite{zhao20193d}  & $88.9$  &   &  \\
            MultiTask~\cite{hassani2019unsupervised}  & $89.1$  &  &  \\
            \hline
            $F_{p}$-CM     & $87.5$  &   &  \\
            $F_{p}$-CM-CV     & $\textbf{89.8}$  &   &  \\
            \hline
        \end{tabular}
        }
    \end{center}
    \label{tab:3D}
\end{table}

Compared to other unsupervised feature learning methods in Table~\ref{tab:3D}, our approach achieves the state-of-the-art accuracy on the ModelNet40 shape recognition task with pre-trained $F_{p}$ and a linear SVM. The performance of $F_{p}$ trained with both cross-modality and cross-view correspondences is $89.8\%$ which is $0.7\%$ higher than the previous state-of-the-art method. Even trained without using any human-annotated labels, the features learned by our network achieve comparable performance as the supervised methods on the ModelNet40 dataset. 

\section{Conclusion}

In this paper, we have proposed a self-supervised learning schema that can jointly learn discriminative 2D and 3D features by using the cross-view and cross-modality correspondences on the 3D point cloud datasets. The learned features from both the 2D image-based network and the 3D point cloud-based graph neural network have been extensively tested across different tasks including multi-view 2D shape recognition, 3D shape recognition, multi-view 2D shape retrieval, 3D shape retrieval, and 3D part-segmentation, showing strong generalization abilities of the learned features. Our results demonstrate a promising direction to learn features by exploiting cross-modality correspondence among different modalities derived from 3D data including mesh, rendered multi-view data, voxel, point cloud, Phong, depth, Silhouette, etc.

\section{Acknowledgement}
This material is partially based upon the work supported by National Science Foundation (NSF) under award number IIS-1400802.

\bibliographystyle{splncs04}
\bibliography{CrossModalView}

\begin{thebibliography}{10}
\providecommand{\url}[1]{\texttt{#1}}
\providecommand{\urlprefix}{URL }
\providecommand{\doi}[1]{https://doi.org/#1}

\bibitem{achlioptas2017learning}
Achlioptas, P., Diamanti, O., Mitliagkas, I., Guibas, L.: Learning
  representations and generative models for 3d point clouds. arXiv preprint
  arXiv:1707.02392  (2017)

\bibitem{behley2019semantickitti}
Behley, J., Garbade, M., Milioto, A., Quenzel, J., Behnke, S., Stachniss, C.,
  Gall, J.: Semantickitti: A dataset for semantic scene understanding of lidar
  sequences. In: Proceedings of the IEEE International Conference on Computer
  Vision. pp. 9297--9307 (2019)

\bibitem{caron2018deep}
Caron, M., Bojanowski, P., Joulin, A., Douze, M.: Deep clustering for
  unsupervised learning of visual features. In: Proceedings of the European
  Conference on Computer Vision (ECCV). pp. 132--149 (2018)

\bibitem{chang2015shapenet}
Chang, A.X., Funkhouser, T., Guibas, L., Hanrahan, P., Huang, Q., Li, Z.,
  Savarese, S., Savva, M., Song, S., Su, H., et~al.: Shapenet: An
  information-rich 3d model repository. arXiv preprint arXiv:1512.03012  (2015)

\bibitem{chen2003visual}
Chen, D.Y., Tian, X.P., Shen, Y.T., Ouhyoung, M.: On visual similarity based 3d
  model retrieval. In: Computer graphics forum. vol.~22, pp. 223--232. Wiley
  Online Library (2003)

\bibitem{deng2009imagenet}
Deng, J., Dong, W., Socher, R., Li, L.J., Li, K., Fei-Fei, L.: Imagenet: A
  large-scale hierarchical image database. In: 2009 IEEE conference on computer
  vision and pattern recognition. pp. 248--255. Ieee (2009)

\bibitem{donahue2014decaf}
Donahue, J., Jia, Y., Vinyals, O., Hoffman, J., Zhang, N., Tzeng, E., Darrell,
  T.: Decaf: A deep convolutional activation feature for generic visual
  recognition. In: International conference on machine learning. pp. 647--655
  (2014)

\bibitem{feat1}
Fang, Y., Xie, J., Dai, G., Wang, M., Zhu, F., Xu, T., Wong, E.: 3d deep shape
  descriptor. In: CVPR. pp. 2319--2328 (2015)

\bibitem{gadelha2018multiresolution}
Gadelha, M., Wang, R., Maji, S.: Multiresolution tree networks for 3d point
  cloud processing. In: Proceedings of the European Conference on Computer
  Vision (ECCV). pp. 103--118 (2018)

\bibitem{gidaris2018unsupervised}
Gidaris, S., Singh, P., Komodakis, N.: Unsupervised representation learning by
  predicting image rotations. arXiv preprint arXiv:1803.07728  (2018)

\bibitem{girdhar2016learning}
Girdhar, R., Fouhey, D.F., Rodriguez, M., Gupta, A.: Learning a predictable and
  generative vector representation for objects. In: European Conference on
  Computer Vision. pp. 484--499. Springer (2016)

\bibitem{goyal2019scaling}
Goyal, P., Mahajan, D., Gupta, A., Misra, I.: Scaling and benchmarking
  self-supervised visual representation learning. In: Proceedings of the IEEE
  International Conference on Computer Vision. pp. 6391--6400 (2019)

\bibitem{feat2}
Guo, K., Zou, D., Chen, X.: 3d mesh labeling via deep convolutional neural
  networks. TOG  \textbf{35}(1), ~3 (2015)

\bibitem{hassani2019unsupervised}
Hassani, K., Haley, M.: Unsupervised multi-task feature learning on point
  clouds. In: Proceedings of the IEEE International Conference on Computer
  Vision. pp. 8160--8171 (2019)

\bibitem{he2016deep}
He, K., Zhang, X., Ren, S., Sun, J.: Deep residual learning for image
  recognition. In: Proceedings of the IEEE conference on computer vision and
  pattern recognition. pp. 770--778 (2016)

\bibitem{hua2018pointwise}
Hua, B.S., Tran, M.K., Yeung, S.K.: Pointwise convolutional neural networks.
  In: Proceedings of the IEEE Conference on Computer Vision and Pattern
  Recognition. pp. 984--993 (2018)

\bibitem{jing2018self}
Jing, L., Tian, Y.: Self-supervised spatiotemporal feature learning by video
  geometric transformations. arXiv preprint arXiv:1811.11387  \textbf{2}(7), ~8
  (2018)

\bibitem{jing2019self}
Jing, L., Tian, Y.: Self-supervised visual feature learning with deep neural
  networks: A survey. arXiv preprint arXiv:1902.06162  (2019)

\bibitem{kay2017kinetics}
Kay, W., Carreira, J., Simonyan, K., Zhang, B., Hillier, C., Vijayanarasimhan,
  S., Viola, F., Green, T., Back, T., Natsev, P., et~al.: The kinetics human
  action video dataset. arXiv preprint arXiv:1705.06950  (2017)

\bibitem{kazhdan2003rotation}
Kazhdan, M., Funkhouser, T., Rusinkiewicz, S.: Rotation invariant spherical
  harmonic representation of 3 d shape descriptors. In: Symposium on geometry
  processing. vol.~6, pp. 156--164 (2003)

\bibitem{kingma2014adam}
Kingma, D.P., Ba, J.: Adam: A method for stochastic optimization. arXiv
  preprint arXiv:1412.6980  (2014)

\bibitem{volum4}
Klokov, R., Lempitsky, V.: Escape from cells: Deep kd-networks for the
  recognition of 3d point cloud models. In: ICCV. pp. 863--872 (2017)

\bibitem{klokov2017escape}
Klokov, R., Lempitsky, V.: Escape from cells: Deep kd-networks for the
  recognition of 3d point cloud models. In: Proceedings of the IEEE
  International Conference on Computer Vision. pp. 863--872 (2017)

\bibitem{kolesnikov2019revisiting}
Kolesnikov, A., Zhai, X., Beyer, L.: Revisiting self-supervised visual
  representation learning. In: Proceedings of the IEEE conference on Computer
  Vision and Pattern Recognition. pp. 1920--1929 (2019)

\bibitem{korbar2018cooperative}
Korbar, B., Tran, D., Torresani, L.: Cooperative learning of audio and video
  models from self-supervised synchronization. In: Advances in Neural
  Information Processing Systems. pp. 7763--7774 (2018)

\bibitem{lang2019pointpillars}
Lang, A.H., Vora, S., Caesar, H., Zhou, L., Yang, J., Beijbom, O.:
  Pointpillars: Fast encoders for object detection from point clouds. In:
  Proceedings of the IEEE Conference on Computer Vision and Pattern
  Recognition. pp. 12697--12705 (2019)

\bibitem{lei2019spherical}
Lei, H., Akhtar, N., Mian, A.: Spherical kernel for efficient graph convolution
  on 3d point clouds. arXiv preprint arXiv:1909.09287  (2019)

\bibitem{li2018point}
Li, C.L., Zaheer, M., Zhang, Y., Poczos, B., Salakhutdinov, R.: Point cloud
  gan. arXiv preprint arXiv:1810.05795  (2018)

\bibitem{li2019deepgcns}
Li, G., Muller, M., Thabet, A., Ghanem, B.: Deepgcns: Can gcns go as deep as
  cnns? In: Proceedings of the IEEE International Conference on Computer
  Vision. pp. 9267--9276 (2019)

\bibitem{maaten2008visualizing}
Maaten, L.v.d., Hinton, G.: Visualizing data using t-sne. Journal of machine
  learning research  \textbf{9}(Nov),  2579--2605 (2008)

\bibitem{volum1}
Maturana, D., Scherer, S.: Voxnet: A 3d convolutional neural network for
  real-time object recognition. In: IROS. pp. 922--928. IEEE (2015)

\bibitem{noroozi2016unsupervised}
Noroozi, M., Favaro, P.: Unsupervised learning of visual representations by
  solving jigsaw puzzles. In: European Conference on Computer Vision. pp.
  69--84. Springer (2016)

\bibitem{noroozi2018boosting}
Noroozi, M., Vinjimoor, A., Favaro, P., Pirsiavash, H.: Boosting
  self-supervised learning via knowledge transfer. In: Proceedings of the IEEE
  Conference on Computer Vision and Pattern Recognition. pp. 9359--9367 (2018)

\bibitem{pathak2017learning}
Pathak, D., Girshick, R., Doll{\'a}r, P., Darrell, T., Hariharan, B.: Learning
  features by watching objects move. In: Proceedings of the IEEE Conference on
  Computer Vision and Pattern Recognition. pp. 2701--2710 (2017)

\bibitem{pathak2016context}
Pathak, D., Krahenbuhl, P., Donahue, J., Darrell, T., Efros, A.A.: Context
  encoders: Feature learning by inpainting. In: Proceedings of the IEEE
  conference on computer vision and pattern recognition. pp. 2536--2544 (2016)

\bibitem{phong1975illumination}
Phong, B.T.: Illumination for computer generated pictures. Communications of
  the ACM  \textbf{18}(6),  311--317 (1975)

\bibitem{qi2017pointnet}
Qi, C.R., Su, H., Mo, K., Guibas, L.J.: Pointnet: Deep learning on point sets
  for 3d classification and segmentation. In: Proceedings of the IEEE
  conference on computer vision and pattern recognition. pp. 652--660 (2017)

\bibitem{volum5}
Qi, C.R., Su, H., Nie{\ss}ner, M., Dai, A., Yan, M., Guibas, L.J.: Volumetric
  and multi-view cnns for object classification on 3d data. In: CVPR. pp.
  5648--5656 (2016)

\bibitem{qi2017pointnet++}
Qi, C.R., Yi, L., Su, H., Guibas, L.J.: Pointnet++: Deep hierarchical feature
  learning on point sets in a metric space. In: Advances in neural information
  processing systems. pp. 5099--5108 (2017)

\bibitem{russakovsky2015imagenet}
Russakovsky, O., Deng, J., Su, H., Krause, J., Satheesh, S., Ma, S., Huang, Z.,
  Karpathy, A., Khosla, A., Bernstein, M., et~al.: Imagenet large scale visual
  recognition challenge. International journal of computer vision
  \textbf{115}(3),  211--252 (2015)

\bibitem{sanchez2013image}
S{\'a}nchez, J., Perronnin, F., Mensink, T., Verbeek, J.: Image classification
  with the fisher vector: Theory and practice. International journal of
  computer vision  \textbf{105}(3),  222--245 (2013)

\bibitem{sauder2019self}
Sauder, J., Sievers, B.: Self-supervised deep learning on point clouds by
  reconstructing space. In: Advances in Neural Information Processing Systems.
  pp. 12942--12952 (2019)

\bibitem{schroff2015facenet}
Schroff, F., Kalenichenko, D., Philbin, J.: Facenet: A unified embedding for
  face recognition and clustering. In: Proceedings of the IEEE conference on
  computer vision and pattern recognition. pp. 815--823 (2015)

\bibitem{sharma2016vconv}
Sharma, A., Grau, O., Fritz, M.: Vconv-dae: Deep volumetric shape learning
  without object labels. In: European Conference on Computer Vision. pp.
  236--250. Springer (2016)

\bibitem{shen2018mining}
Shen, Y., Feng, C., Yang, Y., Tian, D.: Mining point cloud local structures by
  kernel correlation and graph pooling. In: Proceedings of the IEEE conference
  on computer vision and pattern recognition. pp. 4548--4557 (2018)

\bibitem{su2015multi}
Su, H., Maji, S., Kalogerakis, E., Learned-Miller, E.: Multi-view convolutional
  neural networks for 3d shape recognition. In: Proceedings of the IEEE
  international conference on computer vision. pp. 945--953 (2015)

\bibitem{su2018deeper}
Su, J.C., Gadelha, M., Wang, R., Maji, S.: A deeper look at 3d shape
  classifiers. In: Proceedings of the European Conference on Computer Vision
  (ECCV). pp.~0--0 (2018)

\bibitem{sun2018pointgrow}
Sun, Y., Wang, Y., Liu, Z., Siegel, J.E., Sarma, S.E.: Pointgrow:
  Autoregressively learned point cloud generation with self-attention. arXiv
  preprint arXiv:1810.05591  (2018)

\bibitem{volum3}
Tatarchenko, M., Dosovitskiy, A., Brox, T.: Octree generating networks:
  Efficient convolutional architectures for high-resolution 3d outputs. In:
  ICCV. pp. 2088--2096 (2017)

\bibitem{thabet2019mortonnet}
Thabet, A., Alwassel, H., Ghanem, B.: Mortonnet: Self-supervised learning of
  local features in 3d point clouds. arXiv preprint arXiv:1904.00230  (2019)

\bibitem{thomas2019kpconv}
Thomas, H., Qi, C.R., Deschaud, J.E., Marcotegui, B., Goulette, F., Guibas,
  L.J.: Kpconv: Flexible and deformable convolution for point clouds. In:
  Proceedings of the IEEE International Conference on Computer Vision. pp.
  6411--6420 (2019)

\bibitem{wang2018local}
Wang, C., Samari, B., Siddiqi, K.: Local spectral graph convolution for point
  set feature learning. In: Proceedings of the European conference on computer
  vision (ECCV). pp. 52--66 (2018)

\bibitem{wang2019dynamic}
Wang, Y., Sun, Y., Liu, Z., Sarma, S.E., Bronstein, M.M., Solomon, J.M.:
  Dynamic graph cnn for learning on point clouds. ACM Transactions on Graphics
  (TOG)  \textbf{38}(5),  1--12 (2019)

\bibitem{wu2016learning}
Wu, J., Zhang, C., Xue, T., Freeman, B., Tenenbaum, J.: Learning a
  probabilistic latent space of object shapes via 3d generative-adversarial
  modeling. In: Advances in neural information processing systems. pp. 82--90
  (2016)

\bibitem{wu2019pointconv}
Wu, W., Qi, Z., Fuxin, L.: Pointconv: Deep convolutional networks on 3d point
  clouds. In: Proceedings of the IEEE Conference on Computer Vision and Pattern
  Recognition. pp. 9621--9630 (2019)

\bibitem{wu20153d}
Wu, Z., Song, S., Khosla, A., Yu, F., Zhang, L., Tang, X., Xiao, J.: 3d
  shapenets: A deep representation for volumetric shapes. In: Proceedings of
  the IEEE conference on computer vision and pattern recognition. pp.
  1912--1920 (2015)

\bibitem{yang2018foldingnet}
Yang, Y., Feng, C., Shen, Y., Tian, D.: Foldingnet: Point cloud auto-encoder
  via deep grid deformation. In: Proceedings of the IEEE Conference on Computer
  Vision and Pattern Recognition. pp. 206--215 (2018)

\bibitem{zhang2019unsupervised}
Zhang, L., Zhu, Z.: Unsupervised feature learning for point cloud understanding
  by contrasting and clustering using graph convolutional neural networks. In:
  2019 International Conference on 3D Vision (3DV). pp. 395--404. IEEE (2019)

\bibitem{zhao20193d}
Zhao, Y., Birdal, T., Deng, H., Tombari, F.: 3d point capsule networks. In:
  Proceedings of the IEEE Conference on Computer Vision and Pattern
  Recognition. pp. 1009--1018 (2019)

\end{thebibliography}

\end{document}